\documentclass{article}

\usepackage{microtype}
\usepackage{graphicx}
\usepackage{subfigure}
\usepackage{booktabs} 
\usepackage{multirow}
\usepackage{hyperref}
\usepackage{subcaption}  

\usepackage[accepted]{icml2025}

\usepackage{amsmath}
\usepackage{amssymb}
\usepackage{mathtools}
\usepackage{amsthm}

\usepackage[capitalize,noabbrev]{cleveref}

\theoremstyle{plain}

\theoremstyle{definition}

\theoremstyle{remark}

\usepackage[textsize=tiny]{todonotes}

\begin{document}

\twocolumn[
\icmltitle{Predicting Microbial Ontology and Pathogen Risk from Environmental Metadata with Large Language Models}



\icmlsetsymbol{equal}{*}

\begin{icmlauthorlist}
\icmlauthor{Hyunwoo Yoo}{equal,yyy,comp}
\icmlauthor{Gail L. Rosen}{equal,yyy}
\end{icmlauthorlist}

\icmlaffiliation{yyy}{Department of Electrical and Computer Engineering, Drexel University, Philadelphia, United States}
\icmlaffiliation{comp}{MODULABS, Seoul, Korea}

\icmlcorrespondingauthor{Hyunwoo Yoo}{hty23@drexel.edu}
\icmlcorrespondingauthor{Gail L. Rosen}{glr26@drexel.edu}

\icmlkeywords{Machine Learning, ICML}

\vskip 0.3in
]



\printAffiliationsAndNotice{\icmlEqualContribution} 

\begin{abstract}
Traditional machine learning models struggle to generalize in microbiome studies where only metadata is available, especially in small-sample settings or across studies with heterogeneous label formats. In this work, we explore the use of large language models (LLMs) to classify microbial samples into ontology categories such as EMPO 3 and related biological labels, as well as to predict pathogen contamination risk, specifically the presence of \textit{E. Coli}, using environmental metadata alone. We evaluate LLMs such as ChatGPT-4o, Claude 3.7 Sonnet, Grok-3, and LLaMA 4 in zero-shot and few-shot settings, comparing their performance against traditional models like Random Forests across multiple real-world datasets. Our results show that LLMs not only outperform baselines in ontology classification, but also demonstrate strong predictive ability for contamination risk, generalizing across sites and metadata distributions. These findings suggest that LLMs can effectively reason over sparse, heterogeneous biological metadata and offer a promising metadata-only approach for environmental microbiology and biosurveillance applications.
\end{abstract}

\section{Introduction}
\label{Introduction}


Microbiome classification often relies on sequencing-based taxonomic features. However, such data may be missing or excluded due to cost or design constraints, leaving only environmental metadata for analysis. 
Environmental metadata, such as material type, biome, sample type, or collection site, is frequently available even in sparse or small-scale studies. Traditional models struggle to generalize on metadata-only settings. This challenge is exacerbated in small-sample settings or when datasets differ in label expressions or structure. These models are limited by their dependence on exact string matches and lack of semantic understanding, which reduces robustness across domains. Even differential abundance methods often yield inconsistent results across datasets, raising reproducibility concerns \cite{nearing2022microbiome}.

Recent advances in large language models (LLMs) present a new opportunity to reason over natural language metadata. Without requiring fine-tuning or manual feature engineering, LLMs can interpret the semantic meaning of metadata fields and align labels across studies. This makes them particularly effective in zero-shot and few-shot classification settings. We adopt these settings to simulate real-world cases where model fine-tuning is not feasible due to sparse or missing sequencing data.

In this work, we investigate whether LLMs can perform two key tasks using environmental metadata alone: classifying microbial samples into ontology categories such as EMPO 3, and predicting pathogen contamination risk, specifically the presence of \textit{E. Coli}. These tasks reflect both ecological characterization and health-related biosurveillance, illustrating the broad utility of LLMs in biological inference.

We evaluate several state-of-the-art LLMs, including ChatGPT-4o~\cite{openai2023gpt4}, Claude 3.7 Sonnet~\cite{anthropic2025claude3_7}, Grok-3~\cite{grok32025}, and LLaMA 4~\cite{meta2025llama4}, under zero-shot and few-shot settings. Their performance is compared to that of traditional models across multiple real-world datasets. Our results show that LLMs not only outperform baselines in ontology classification but also demonstrate strong predictive accuracy for \textit{E. Coli} contamination across varied metadata distributions. These findings highlight the potential of LLMs as metadata-only predictors in microbiome research and environmental health monitoring.

\begin{table*}[t]

\vskip 0.15in
\begin{center}
\begin{small}
\begin{sc}
\begin{tabular}{lcccc}
\toprule
\textbf{Model} & \textbf{Accuracy} & \textbf{Macro Prec.} & \textbf{Macro Rec.} & \textbf{Macro F1} \\
\midrule
ChatGPT-4o (ZS)     & 0.96 & 0.69 & 0.75 & 0.71 \\
Claude 3.7 (ZS)     & 0.85 & 0.80 & 0.83 & 0.71 \\
Grok-3 (ZS)         & 0.96 & 0.69 & 0.75 & 0.71 \\
LLaMA 4 (ZS)        & 0.59 & 0.44 & 0.40 & 0.40 \\
\midrule
ChatGPT-4o (FS)     & 0.96 & 0.69 & 0.75 & 0.71 \\
Claude 3.7 (FS)     & 0.96 & 0.69 & 0.75 & 0.71 \\
Grok-3 (FS)         & 0.96 & 0.69 & 0.75 & 0.71 \\
LLaMA 4 (FS)        & 1.00 & 1.00 & 1.00 & 1.00 \\
\midrule
Random Forest       & 0.11 & 0.03 & 0.25 & 0.05 \\
\bottomrule
\end{tabular}
\end{sc}
\end{small}
\end{center}
\vskip -0.1in

\caption{
Zero-shot and Few-shot inference on Study 15573. In few-shot inference, LLMs receive support examples from Study 1728 (a different domain) as prompt context, enabling cross-study generalization evaluation. LLMs are frozen (no parameter updates), and only Random Forest is trained on Study 1728.
}
\label{table-zeroshot-fewshot-15573}
\end{table*}

\section{Related Work}

Traditional microbiome classification relies heavily on taxonomic features derived from high-throughput sequencing data, such as 16S rRNA gene or shotgun metagenomic profiles~\cite{knights2011supervised, pasolli2016machine}. These methods offer accurate microbial characterization but require sequencing resources and may not be feasible in low-cost or metadata-only settings.

In contrast, studies that rely exclusively on environmental metadata for microbiome prediction remain scarce. While there has been growing interest in metadata curation and standardization~\cite{vangay2021microbiome}, and recent work highlights reproducibility issues in reusing metadata from 16S studies \cite{kang2021reprocessing}, yet few models directly leverage such metadata for ontology or pathogen risk classification.

Recent advancements in large language models (LLMs) have enabled strong zero-shot and few-shot reasoning capabilities across a range of tasks~\cite{brown2020language}. In the biomedical domain, LLMs have been used to encode clinical knowledge~\cite{singhal2023large} and integrate multimodal signals such as images and text for medical prediction tasks~\cite{zhang2024generalist}. However, their application to microbiome-related classification using only sparse metadata remains largely unexplored. Moreover, traditional models are highly sensitive to normalization and compositional characteristics of microbiome data, which limits robustness across datasets \cite{weiss2017normalization}.

Our work bridges this gap by applying foundation models to infer microbial ontology and pathogen contamination risk solely from environmental metadata, demonstrating effective generalization across studies without access to genomic sequences.

\section{Method}

We use pre-trained large language models (LLMs), including ChatGPT-4o, Claude 3.7 Sonnet, Grok-3, and LLaMA 4, to perform microbiome-related classification and regression tasks using environmental metadata as input. Specifically, we investigate two types of inference problems: (1) classification of microbial samples into standardized ontology categories and related biological labels, and (2) prediction of pathogen contamination risk based on contextual environmental features.

\begin{table}[h!]
\centering
\begin{scriptsize}
\begin{tabular}{lcccc}
\toprule
\textbf{Model} & \textbf{Acc.} & \textbf{Prec.} & \textbf{Rec.} & \textbf{F1} \\
\midrule
Random Forest & 0.47 & 0.33 & 0.27 & 0.30 \\
ChatGPT-4o (ZS) & 1.00 & 1.00 & 1.00 & 1.00 \\
Claude 3.7 (ZS) & 1.00 & 1.00 & 1.00 & 1.00 \\
Grok-3 (ZS) & 1.00 & 1.00 & 1.00 & 1.00 \\
\bottomrule
\end{tabular}
\end{scriptsize}
\vskip -0.05in
\caption{Zero-shot inference on Study 1728. Only Random Forest is trained on Study 15573; LLMs are zero-shot prompted.}
\label{table-zeroshot-1728}
\end{table}

\subsection{Prompting Strategy for LLM-Based Inference}

All models are accessed in a frozen setting without fine-tuning. We construct prompts in natural English using structured metadata fields as context. Each prompt queries the model about the most likely label or value associated with a given sample.

\paragraph{Zero-shot classification.}  
In zero-shot settings, a sample \( x \) is represented by its metadata fields (e.g., \textit{env material}, \textit{sample type}, \textit{scientific name}, \textit{geo loc name}). We format a multiple-choice prompt that asks whether sample \( x \) corresponds to a candidate label \( y \in \mathcal{Y} \). The model selects the label with highest likelihood:

\[
\hat{y} = \arg\max_{y \in \mathcal{Y}} \text{LLM}_\theta(P(x, y))
\]

\paragraph{Few-shot classification.}  
In few-shot settings, the prompt includes a small number of labeled support examples \( \mathcal{D}_{\text{support}} = \{(x_i, y_i)\}_{i=1}^k \), prepended to the test sample \( x' \). The model selects the most probable label as:

\[
\hat{y} = \arg\max_{y \in \mathcal{Y}} \text{LLM}_\theta(P(\mathcal{D}_{\text{support}}, x', y))
\]

\begin{table*}[h!]

\vskip 0.15in
\begin{center}
\begin{small}
\begin{sc}
\begin{tabular}{lcccc}
\toprule
\textbf{Model} & \textbf{Accuracy} & \textbf{Macro Prec.} & \textbf{Macro Rec.} & \textbf{Macro F1} \\
\midrule
ChatGPT-4o (ZS)        & 0.7500 & 0.7084 & 0.7210 & 0.7135 \\
Claude 4sonet (ZS)     & 0.8036 & 0.7841 & 0.7262 & 0.7441 \\
Grok-3 (ZS)            & 0.7679 & 0.7366 & 0.7670 & 0.7443 \\
LLaMA 4 (ZS)           & 0.7143 & 0.6621 & 0.6621 & 0.6621 \\
Gemini 2.5flash (ZS)   & 0.7857 & 0.7480 & 0.7300 & 0.7375 \\
\midrule
ChatGPT-4o (FS)        & 0.8214 & 0.8202 & 0.7391 & 0.7619 \\
Claude 4sonet (FS)     & 0.7500 & 0.7044 & 0.7044 & 0.7044 \\
Grok-3 (FS)            & 0.8036 & 0.8022 & 0.7097 & 0.7316 \\
LLaMA 4 (FS)           & 0.7500 & 0.7024 & 0.6878 & 0.6937 \\
Gemini 2.5flash (FS)   & 0.7857 & 0.7825 & 0.6802 & 0.6995 \\
\bottomrule
\end{tabular}
\end{sc}
\end{small}
\end{center}
\vskip -0.1in

\caption{
Zero-shot and Few-shot \textit{E. Coli} level binary prediction on the 2005 Huntington Beach dataset. In the few-shot setting, LLMs receive support examples from 2006 samples, enabling evaluation of cross-year generalization across temporally shifted metadata distributions.
}
\label{table-zeroshot-fewshot-2005}
\end{table*}

To evaluate cross-study generalization, the support examples are drawn from a different dataset than the test set (e.g., Study 1728 as support for Study 15573 and vice versa), and no model parameters are updated. Similarly, in the \textit{E. Coli} contamination prediction task, we evaluate cross-year generalization by using 2005 Huntington Beach data for testing and 2006 samples as support examples in few-shot prompting.

\paragraph{Binary contamination prediction.}  
For \textit{E. Coli} contamination risk, we cast the task as binary classification using a prompt that asks whether the microbial contamination level exceeds a threshold such as 126 CFU/100mL as per EPA guidelines\cite{epa2012rwqc, epa2021ecoli}. We evaluate LLM responses against ground-truth labels derived from measured \textit{E. Coli} levels.

\paragraph{Regression prediction.}  
We additionally evaluate whether LLMs can perform numeric regression by prompting for the predicted concentration of \textit{E. Coli} based on a sample's environmental metadata. The model's free-text numeric output is extracted and compared to ground-truth concentrations using standard regression metrics such as MAE and \( R^2 \).

\subsection{Prompt Format and Standardization}

All prompts follow a fixed structure, include consistent field ordering, and avoid ambiguous phrasing. We use prompt templates that include direct natural language questions (e.g., "Given the following metadata, what is the most likely EMPO 3 category for this sample?"). For regression tasks, prompts explicitly ask for a numerical estimate (e.g., "Estimate the \textit{E. Coli} concentration (in CFU/100mL) based on the following metadata.").
Examples of all prompt templates are included in Appendix~\ref{app:prompt}.

\section{Experiments}

We evaluate large language models (LLMs) on two tasks using only environmental metadata: (1) microbial context classification and (2) \textit{E. Coli} contamination risk prediction. We consider zero-shot and few-shot prompting with no fine-tuning, and compare LLMs to traditional machine learning models such as Random Forest and XGBoost. Details of environmental metadata are in Appendix~\ref{app:dataset_details}

\begin{table}[h!]
\centering
\begin{scriptsize}
\begin{tabular}{lcccc}
\toprule
\textbf{Model} & \textbf{Acc.} & \textbf{Prec.} & \textbf{Rec.} & \textbf{F1} \\
\midrule
ChatGPT-4o (ZS)      & 0.6721 & 0.6482 & 0.6424 & 0.6445 \\
Claude 4sonet (ZS)   & 0.7049 & 0.6872 & 0.6602 & 0.6652 \\
Grok-3 (ZS)          & 0.7049 & 0.6889 & 0.6945 & 0.6909 \\
LLaMA 4 (ZS)         & 0.7377 & 0.7398 & 0.7551 & 0.7342 \\
Gemini 2.5flash (ZS)      &   NA   &   NA   &   NA   &   NA   \\
\bottomrule
\end{tabular}
\end{scriptsize}
\vskip -0.05in
\caption{Zero-shot \textit{E. Coli} binary prediction on 2006 Huntington Beach data.}
\label{table-zeroshot-2006}
\end{table}

\subsection{Microbial Context Classification}

We begin with EMPO 3 classification using metadata fields such as `env material` and `sample type` in study 1728 and 15573\cite{hewson2022daspc, baum1728}. 
Table~\ref{table-zeroshot-1728} shows that all LLMs achieve perfect accuracy on Study 1728 in zero-shot prompting, while Random Forest trained on Study 15573 fails (47\%). This highlights LLMs' ability to generalize across studies using metadata alone.
On Study 15573, which contains a more diverse set of environments and labels, zero-shot performance remains high for ChatGPT-4o and Grok-3 (96\% accuracy), while Claude 3.7 trails slightly (85\%) and LLaMA 4 struggles at 59\% (Table~\ref{table-zeroshot-fewshot-15573}). This indicates LLMs' ability to perform semantic alignment between metadata and ontology labels across domains. When given few-shot prompts using Study 1728 samples (Table~\ref{table-zeroshot-fewshot-15573}), all models improve or maintain strong performance, with LLaMA 4 jumping to 100\% accuracy. Additional results on `sample\_type` and `scientific\_name` classification appear in Tables~\ref{table-zeroshot-sample-type-15573}–\ref{table-zeroshot-scientific-name-no-sample-type-15573} in Appendix~\ref{app:additional_results}.

\subsection{\textit{E. Coli} Contamination Prediction}

\begin{table}[h!]
\centering
\begin{scriptsize}
\begin{tabular}{lcccc}
\toprule
\textbf{Model} & \textbf{Acc.} & \textbf{Prec.} & \textbf{Rec.} & \textbf{F1} \\
\midrule
ChatGPT-4o (FS)      &  NA   &  NA   &  NA   &  NA   \\
Claude 4sonet (FS)   & 0.6885 & 0.6866 & 0.6985 & 0.6831 \\
Grok-3 (FS)          & 0.7705 & 0.7567 & 0.7643 & 0.7596 \\
LLaMA 4 (FS)         &  NA   &  NA   &  NA   &  NA   \\
Gemini 2.5flash (FS)      & 0.8033 & 0.7906 & 0.7906 & 0.7906 \\
\bottomrule
\end{tabular}
\end{scriptsize}
\vskip -0.05in
\caption{Few-shot \textit{E. Coli} binary prediction on 2006 Huntington Beach data.}
\label{table-fewshot-2006}
\end{table}

We next evaluate binary classification of \textit{E. Coli} risk (above/below regulatory threshold). As shown in the table~\ref{table-zeroshot-fewshot-15573} on 2005 Huntington Beach data, zero-shot prompting shows strong performance. For instance, Claude 4 Sonnet reaches 80.4\% accuracy and 0.7441 macro F1, followed closely by Grok-3 and Gemini 2.5Flash. Few-shot prompting further boosts ChatGPT-4o to 82.1\% accuracy and 0.7619 F1, outperforming all others. This setup mirrors realistic scenarios where prior year data is used to inform current year risk, enabling cross-year generalization without sequence data.

On 2006 data\cite{francy2021ecoli}, LLaMA 4 surprisingly leads in zero-shot performance (73.8\% accuracy), while Claude and Grok-3 also perform well (Table~\ref{table-zeroshot-2006}). Gemini fails to return predictions. In the few-shot setting (Table~\ref{table-fewshot-2006}), Gemini achieves the highest overall score (80.3\% accuracy), followed by Grok-3 and Claude. This suggests LLMs generalize reasonably well across yearly shifts in metadata and environment.




\begin{figure}[h!]
    \centering
    \begin{minipage}{0.238\textwidth}
        \centering
        \includegraphics[width=\textwidth]{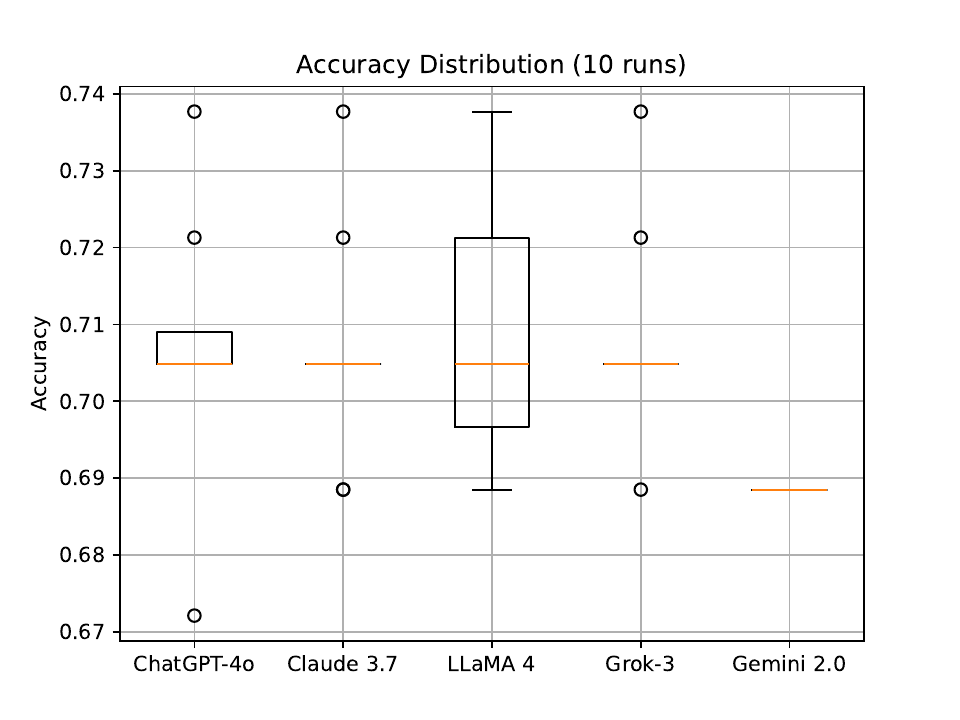}
        \\ \textbf{(a)} Accuracy distribution
    \end{minipage}
    \hfill
    \begin{minipage}{0.238\textwidth}
        \centering
        \includegraphics[width=\textwidth]{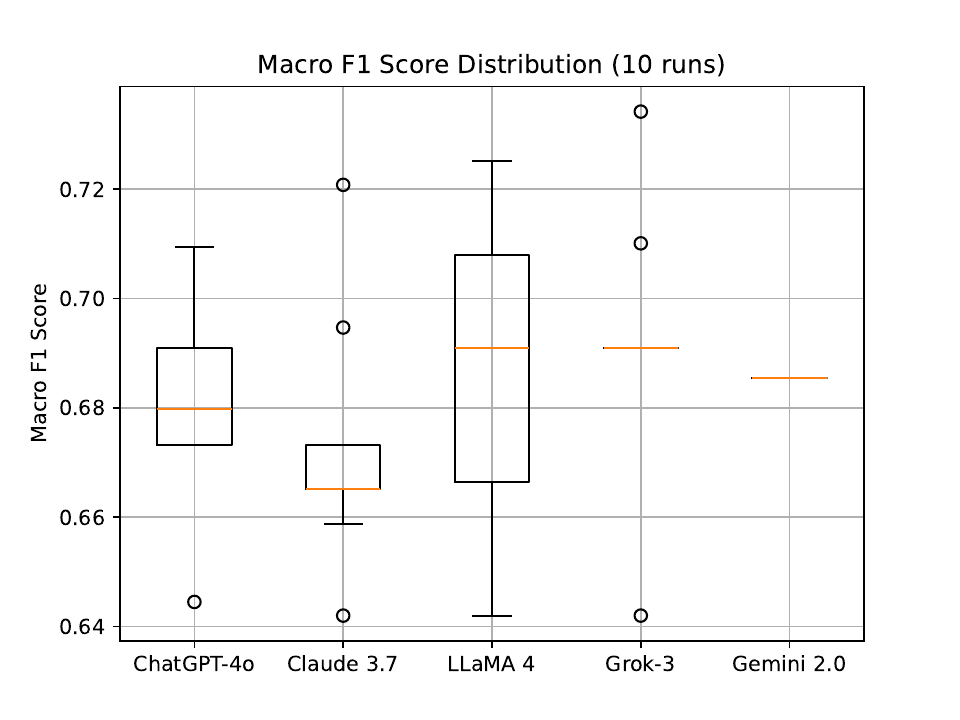}
        \\ \textbf{(b)} F1 Score distribution
    \end{minipage}
    \caption{
        Accuracy and F1 Score distributions across 10 repeated trials for each LLM in \textit{E. Coli} level binary prediction using the 2005 Huntington Beach dataset.
    }
    \label{fig:accuracy_f1_boxplot}
\end{figure}

\subsection{Robustness and Regression Results}

We repeat binary classification over 10 random trials on the 2005 dataset. Figure~\ref{fig:accuracy_f1_boxplot} and Table~\ref{tab:llm_stats} in Appendix~\ref{app:additional_results} show that Grok-3 and Claude 4 Sonnet exhibit the most stable performance. LLaMA 4 displays high variance due to limited valid outputs (N=3), and Gemini's single-run result suggests limited robustness.

For numeric prediction of \textit{E. Coli} concentration, zero-shot regression fails for most LLMs (Table~\ref{tab:llm_regression_zeroshot}). Claude 4 Sonnet returns highly unstable estimates with \( R^2 = -7.71 \). In contrast, in the few-shot setting (Table~\ref{tab:llm_regression_fewshot}), Claude achieves \( R^2 = 0.3946 \), outperforming all traditional models, including Random Forest (0.3261). However, other LLMs such as ChatGPT-4o and Grok-3 still underperform, and variance remains high. This suggests that while LLMs are effective for classification, they are still unreliable for quantitative microbial estimation.

\section{Results and Discussion}
\label{Results}

LLMs consistently outperform traditional models in ontology classification, even when tested on out-of-distribution samples. Cross-study generalization is especially strong for ChatGPT-4o and Grok-3.

While zero-shot prompting already achieves strong binary classification of \textit{E. Coli} presence, few-shot examples further enhance accuracy and F1. Claude and Grok-3 are the most robust models across years.

Despite some success from Claude 4 Sonnet in few-shot settings, LLMs are not yet competitive with traditional regressors for numeric estimation. Prediction variance and output formatting remain issues.


\begin{table}[t]
\centering
\begin{scriptsize}
\begin{tabular}{lcccc}
\toprule
\textbf{Model} & \textbf{MAE} & \textbf{RMSE} & \textbf{MSE} & \textbf{R\textsuperscript{2}} \\
\midrule
ChatGPT-4o      & 287.56 & 577.70  & 333733.36 & -0.0673 \\
Claude 4sonet   & 347.20 & 1650.71 & 2724857.43 & -7.7145 \\
LLaMA 4         & N/A    & N/A     & N/A       & N/A     \\
Grok-3          & 261.32 & 564.57  & 318737.63 & -0.0194 \\
Gemini 2.5flash      & N/A    & N/A     & N/A       & N/A     \\
\bottomrule
\end{tabular}
\end{scriptsize}
\vskip -0.05in
\caption{Zero-shot regression performance on 2006 Huntington beach \textit{E. Coli} data (MAE, RMSE, MSE, $R^2$).}
\label{tab:llm_regression_zeroshot}
\end{table}

\begin{table}[t]
\centering
\begin{scriptsize}
\begin{tabular}{lcccc}
\toprule
\textbf{Model} & \textbf{MAE} & \textbf{RMSE} & \textbf{MSE} & \textbf{R\textsuperscript{2}} \\
\midrule
ChatGPT-4o          & 259.14 & 612.07 & 374632.53 & -0.1981 \\
Claude 4sonet       & 190.96 & 435.10 & 189310.24 & 0.3946 \\
LLaMA 4             &  N/A   &  N/A   &   N/A     &  N/A   \\
Grok-3              & 190.67 & 493.97 & 244002.06 & 0.2196 \\
Gemini 2.5flash          & 232.34 & 478.71 & 229159.20 & 0.2671 \\
\midrule
Random Forest       & 224.69 & 459.03 & 210709.20 & 0.3261 \\
XGBoost             & 251.89 & 542.17 & 293954.79 & 0.0599 \\
Logistic Reg.       & 220.76 & 524.28 & 274872.34 & 0.1209 \\
\bottomrule
\end{tabular}
\end{scriptsize}
\vskip -0.05in
\caption{Few-shot regression performance on 2006 Huntington beach \textit{E. Coli} data (MAE, RMSE, MSE, $R^2$).}
\label{tab:llm_regression_fewshot}
\end{table}

\section{Conclusion}

We show that large language models (LLMs) can accurately classify microbial ontology labels and predict pathogen contamination risk using only environmental metadata. Across diverse datasets and tasks, LLMs demonstrate strong zero-shot and few-shot performance, often outperforming traditional models without requiring fine-tuning. These results highlight the capacity of LLMs to semantically reason over heterogeneous metadata fields, enabling generalization across studies and label variations. While LLM-based regression remains less reliable, classification results suggest that foundation models offer a promising sequence-free approach for microbiome analysis and environmental biosurveillance.

\section*{Acknowledgements}
This work was supported in part by the National Science Foundation (NSF) under Grant Number 2107108 and Brian Impact Foundation, a non-profit organization dedicated to the advancement of science and technology for all.

\newpage
\bibliography{example_paper}
\bibliographystyle{icml2025}

\newpage
\appendix
\onecolumn
\section{Dataset Details}
\label{app:dataset_details}

Study 1728 (Baum asphalt 1st submission)\cite{baum1728} involved bacterial metagenomic sequencing of samples collected from various asphalt sites as well as nearby water and soil locations. Additionally, three samples (2H, 2L, and 2O) were cultured in LB broth before being included in the overall analysis. This study comprises a total of 17 samples, categorized into two EMPO 3 labels: 10 samples as Solid (non-saline) and 7 samples as Aqueous (non-saline).

The metadata for this study includes five primary features: env\_material, env\_biome, env\_feature, sample\_type, and scientific\_name. The feature values are relatively simple and consistent. For env\_material, the categories include water, soil, and anthropogenic environmental material. The env\_biome is exclusively desert biome. All samples are associated with the env\_feature road. The sample\_type includes water filter, water and LB broth, soil, and asphalt. The scientific\_name consists of freshwater metagenome, soil metagenome, and outdoor metagenome. This study does not include geographic information such as geo\_loc\_name, and the overall structure of metadata is constrained and minimal.

Study 15573 (Detection of the Diadema antillarum scuticociliatosis Philaster clade on sympatric metazoa, plankton, and abiotic surfaces and assessment for its potential reemergence)~\cite{hewson2022daspc} is an investigation into the distribution and persistence of a scuticociliate pathogen (DaScPc) that caused mass mortality of long-spined sea urchins throughout the eastern Caribbean in 2022. A total of 27 samples were analyzed, collected from coral species, turf algae, and sponges, particularly near the original outbreak site. The EMPO 3 label distribution includes 17 samples as Animal (saline), 6 as Plant (saline), 3 as Solid (non-saline), and 1 as Aqueous (saline), indicating greater label diversity compared to Study 1728.

The metadata for Study 15573 contains six features: env\_material, env\_biome, env\_feature, sample\_type, scientific\_name, and geo\_loc\_name. The env\_material includes organic material and anthropogenic environmental material. The env\_biome consists of marine biome and urban biome. The env\_feature spans coral reef, animal-associated habitat, plant-associated habitat, anthropogenic environmental feature, and research facility. The sample\_type includes categories such as Turf Algae, coral, hydrozoans, sponge, Boat Hull, Mangrove Leaf, and control swab. The scientific\_name categories include algae metagenome, coral metagenome, sponge metagenome, and plant metagenome. The geo\_loc\_name feature provides geographic labels such as US Virgin Islands and Aruba, offering richer contextual information.

Overall, Study 15573 features more complex metadata with greater categorical diversity across multiple fields and includes detailed geographic context. In comparison, Study 1728 is more constrained in terms of both label space and feature variety. These differences contribute to substantial domain shift, which presents a challenge for conventional machine learning models trained on one study and evaluated on the other.

We also utilize a subset of the dataset released by the U.S. Geological Survey as part of the Great Lakes NowCast project~\cite{francy2021ecoli}, specifically the calibration data collected at Huntington Beach, Ohio, during the 2005 and 2006 recreational seasons.
The dataset contains seven variables. Date indicates the sampling date. EcoliAve\_CFU represents the average concentration of \textit{Escherichia coli} (\textit{E. Coli}), measured in colony-forming units (CFU). Lake\_Temp\_C denotes the lake water temperature in degrees Celsius, while Lake\_Turb\_NTRU captures the turbidity of the lake, measured in Nephelometric Turbidity Units (NTU). WaveHt\_Ft indicates the height of waves in feet at the time of sampling. LL\_PreDay refers to the change in lake level compared to the previous day. Lastly, AirportRain48W\_in measures the cumulative rainfall over the past 48 hours as recorded at a nearby airport, expressed in inches.


\section{Addtional Results}
\label{app:additional_results}

\begin{table*}[h!]

\vskip 0.15in
\begin{center}
\begin{small}
\begin{sc}
\begin{tabular}{lcccc}
\toprule
\textbf{Model} & \textbf{Accuracy} & \textbf{Macro Prec.} & \textbf{Macro Rec.} & \textbf{Macro F1} \\
\midrule
ChatGPT-4o (ZS)     & 1.0 & 1.0 & 1.0 & 1.0 \\
Claude 3.7 (ZS)     & 1.0 & 1.0 & 1.0 & 1.0 \\
Grok-3 (ZS)         & 1.0 & 1.0 & 1.0 & 1.0 \\
LLaMA 4 (ZS)        & 1.0 & 1.0 & 1.0 & 1.0 \\
\bottomrule
\end{tabular}
\end{sc}
\end{small}
\end{center}
\vskip -0.1in

\caption{Zero-shot sample\_type prediction on Study 15573.}
\label{table-zeroshot-sample-type-15573}

\end{table*}

\begin{table*}[h!]

\vskip 0.15in
\begin{center}
\begin{small}
\begin{sc}
\begin{tabular}{lcccc}
\toprule
\textbf{Model} & \textbf{Accuracy} & \textbf{Macro Prec.} & \textbf{Macro Rec.} & \textbf{Macro F1} \\
\midrule
ChatGPT-4o (ZS)     & 1.0 & 1.0 & 1.0 & 1.0 \\
Claude 3.7 (ZS)     & 1.0 & 1.0 & 1.0 & 1.0 \\
Grok-3 (ZS)         & 1.0 & 1.0 & 1.0 & 1.0 \\
LLaMA 4 (ZS)        & 1.0 & 1.0 & 1.0 & 1.0 \\
\bottomrule
\end{tabular}
\end{sc}
\end{small}
\end{center}
\vskip -0.1in

\caption{Zero-shot scientific\_name prediction on Study 15573.}
\label{table-zeroshot-scientific-name-15573}

\end{table*}

\begin{table*}[h!]

\vskip 0.15in
\begin{center}
\begin{small}
\begin{sc}
\begin{tabular}{lcccc}
\toprule
\textbf{Model} & \textbf{Accuracy} & \textbf{Macro Prec.} & \textbf{Macro Rec.} & \textbf{Macro F1} \\
\midrule
ChatGPT-4o (ZS)     & 0.4074 & 0.3824 & 0.5000 & 0.4091 \\
Claude 3.7 (ZS)     & 0.4074 & 0.3824 & 0.5000 & 0.4091 \\
Grok-3 (ZS)         & 0.4074 & 0.3824 & 0.5000 & 0.4091 \\
LLaMA 4 (ZS)        & 0.4074 & 0.3824 & 0.5000 & 0.4091 \\
\bottomrule
\end{tabular}
\end{sc}
\end{small}
\end{center}
\vskip -0.1in

\caption{Zero-shot scientific\_name prediction on Study 15573 without sample\_type feature.}
\label{table-zeroshot-scientific-name-no-sample-type-15573}

\end{table*}

We present additional zero-shot classification results on Study 15573 to further evaluate the robustness of LLMs under different label and feature configurations.

\paragraph{Sample type prediction.}
As shown in Table~\ref{table-zeroshot-sample-type-15573}, all LLMs achieved perfect accuracy when predicting the sample\_type field using available metadata. This confirms that even simple categorical labels can be reliably inferred by LLMs in zero-shot settings without fine-tuning.

\paragraph{Scientific name prediction.}
Table~\ref{table-zeroshot-scientific-name-15573} reports results for predicting scientific\_name labels. Again, all LLMs achieved perfect accuracy, indicating strong semantic reasoning over metadata features that are biologically descriptive.

\paragraph{Impact of removing sample\_type.}
To assess the importance of the features, we removed the sample\_type field and repeated the scientific\_name classification. As seen in Table~\ref{table-zeroshot-scientific-name-no-sample-type-15573}, accuracy drops significantly to 40.7\% for all models, with macro F1 around 0.41. This suggests that sample\_type is a critical contextual cue for correctly identifying scientific labels and that LLMs are sensitive to the availability of relevant metadata features.

\begin{table*}[h!]
\vskip 0.15in
\begin{center}
\begin{small}
\begin{sc}
\begin{tabular}{llccccc}
\toprule
\textbf{Model} & \textbf{Metric} & \textbf{Mean} & \textbf{Std} & \textbf{Min} & \textbf{Max} & \textbf{Valid N} \\
\midrule
\multirow{4}{*}{ChatGPT-4o}
  & Accuracy  &0.7070  &0.0173  &0.6721  &0.7377  &8  \\
  & Precision &0.6892  &0.0204  &0.6482  &0.7231  &8  \\
  & Recall    &0.6779  &0.0182  &0.6424  &0.7037  &8  \\
  & F1        &0.6802  &0.0175  &0.6445  &0.7095  &8  \\
\midrule
\multirow{4}{*}{Claude 4 sonet}
  & Accuracy  &0.7067  &0.0143  &0.6885  &0.7377  &9  \\
  & Precision &0.6877  &0.0157  &0.6659  &0.7208  &9  \\
  & Recall    &0.6693  &0.0223  &0.6384  &0.7208  &9  \\
  & F1        &0.6731  &0.0213  &0.6420  &0.7208  &9  \\
\midrule
\multirow{4}{*}{LLaMA 4}
  & Accuracy  &0.7104  &0.0205  &0.6885  &0.7377  &3  \\
  & Precision &0.6933  &0.0225  &0.6681  &0.7228  &3  \\
  & Recall    &0.6874  &0.0375  &0.6384  &0.7294  &3  \\
  & F1        &0.6860  &0.0341  &0.6420  &0.7252  &3  \\
\midrule
\multirow{4}{*}{Grok-3}
  & Accuracy  &0.7085  &0.0129  &0.6885  &0.7377  &9  \\
  & Precision &0.6943  &0.0186  &0.6681  &0.7398  &9  \\
  & Recall    &0.6974  &0.0283  &0.6384  &0.7551  &9  \\
  & F1        &0.6924  &0.0226  &0.6420  &0.7342  &9  \\
\midrule
\multirow{4}{*}{Gemini 2.5 flash}
  & Accuracy  &0.6885  &0.0000  &0.6885  &0.6885  &1  \\
  & Precision &0.6950  &0.0000  &0.6950  &0.6950  &1  \\
  & Recall    &0.7071  &0.0000  &0.7071  &0.7071  &1  \\
  & F1        &0.6855  &0.0000  &0.6855  &0.6855  &1  \\
\bottomrule
\end{tabular}
\end{sc}
\end{small}
\end{center}
\vskip -0.1in
\caption{
Detailed statistics of LLM performance across 10 repeated runs (or fewer if response failures occurred) in Zero-shot \textit{E. Coli} level binary prediction based on the Data collected in 2006 from Huntington Beach, Ohio. We report the mean, standard deviation, minimum, and maximum values, as well as the number of valid runs (Valid N) for each metric.
}
\label{tab:llm_stats}
\end{table*}

To assess the robustness of each LLM, we repeated the zero-shot \textit{E. Coli} classification task 10 times using the 2006 Huntington Beach dataset. Table~\ref{tab:llm_stats} summarizes the mean, standard deviation, and range of evaluation metrics across repeated runs. Notably, Claude 4 Sonnet and Grok-3 exhibited the most consistent performance, while models like LLaMA 4 and Gemini 2.5 Flash showed higher variance or failed to return valid predictions in multiple runs. These findings highlight the varying stability and reliability of LLMs in structured binary classification tasks.

\section{Prompt}
\label{app:prompt}

\subsection{Zero-shot EMPO\_3 classification prompt example}
{\tiny
\begin{verbatim}
Could you predict the "empo_3" values below(which is now in '?') with Animal (saline) or
Plant (saline) or Solid (non-saline) or Aqueous (saline) ?


env_material env_biome env_feature sample_type scientific_name geo_loc_name empo_3
0 organic material marine biome coral reef Turf Algae algae metagenome US Virgin Islands ?
1 organic material marine biome animal-associated habitat coral coral metagenome Aruba ?
2 organic material marine biome animal-associated habitat coral coral metagenome US Virgin Islands ?
3 organic material marine biome plant-associated habitat Mangrove Leaf plant metagenome US Virgin Islands ?
4 organic material marine biome coral reef Turf Algae algae metagenome US Virgin Islands ?
5 organic material marine biome animal-associated habitat coral coral metagenome Aruba ?
6 organic material marine biome animal-associated habitat hydrozoans hydrozoan metagenome Aruba ?
7 organic material marine biome animal-associated habitat coral coral metagenome US Virgin Islands ?
8 organic material marine biome animal-associated habitat sponge sponge metagenome US Virgin Islands ?
9 organic material marine biome animal-associated habitat sponge sponge metagenome US Virgin Islands ?
10 organic material marine biome animal-associated habitat coral coral metagenome Aruba ?
11 organic material marine biome coral reef Turf Algae algae metagenome US Virgin Islands ?
12 organic material marine biome animal-associated habitat sponge sponge metagenome US Virgin Islands ?
13 organic material marine biome animal-associated habitat coral coral metagenome US Virgin Islands ?
14 organic material marine biome anthrogenic environmental feature Boat Hull metagenome US Virgin Islands ?
15 organic material marine biome animal-associated habitat sponge sponge metagenome US Virgin Islands ?
16 organic material marine biome animal-associated habitat coral coral metagenome US Virgin Islands ?
17 organic material marine biome animal-associated habitat coral coral metagenome US Virgin Islands ?
18 anthropogenic environmental material urban biome research facility control swab metagenome US Virgin Islands ?
19 anthropogenic environmental material urban biome research facility control swab metagenome US Virgin Islands ?
20 anthropogenic environmental material urban biome research facility control swab metagenome US Virgin Islands ?
21 organic material marine biome plant-associated habitat Mangrove Leaf plant metagenome US Virgin Islands ?
22 organic material marine biome animal-associated habitat coral coral metagenome US Virgin Islands ?
23 organic material marine biome animal-associated habitat coral coral metagenome US Virgin Islands ?
24 organic material marine biome coral reef Turf Algae algae metagenome Aruba ?
25 organic material marine biome animal-associated habitat coral coral metagenome Aruba ?
26 organic material marine biome animal-associated habitat sponge sponge metagenome US Virgin Islands ?


\end{verbatim}
}

\subsection{Few-shot EMPO\_3 classification prompt example}
{\tiny
\begin{verbatim}

Based on this study:

env_material env_biome env_feature sample_type scientific_name empo_3
0 water desert biome road water filter freshwater metagenome Aqueous (non-saline)
1 water desert biome road water filter freshwater metagenome Aqueous (non-saline)
2 soil desert biome road soil soil metagenome Solid (non-saline)
3 anthropogenic environmental material desert biome road asphalt outdoor metagenome Solid (non-saline)
4 anthropogenic environmental material desert biome road asphalt outdoor metagenome Solid (non-saline)
5 anthropogenic environmental material desert biome road asphalt outdoor metagenome Solid (non-saline)
6 anthropogenic environmental material desert biome road asphalt outdoor metagenome Solid (non-saline)
7 water desert biome road water and LB broth freshwater metagenome Aqueous (non-saline)
8 soil desert biome road soil soil metagenome Solid (non-saline)
9 water desert biome road water and LB broth freshwater metagenome Aqueous (non-saline)
10 anthropogenic environmental material desert biome road asphalt outdoor metagenome Solid (non-saline)
11 anthropogenic environmental material desert biome road asphalt outdoor metagenome Solid (non-saline)
12 water desert biome road water and LB broth freshwater metagenome Aqueous (non-saline)
13 anthropogenic environmental material desert biome road asphalt outdoor metagenome Solid (non-saline)
14 anthropogenic environmental material desert biome road asphalt outdoor metagenome Solid (non-saline)
15 water desert biome road water filter freshwater metagenome Aqueous (non-saline)
16 water desert biome road water filter freshwater metagenome Aqueous (non-saline)


env_material env_biome env_feature sample_type scientific_name geo_loc_name empo_3
0 organic material marine biome coral reef Turf Algae algae metagenome US Virgin Islands ?
1 organic material marine biome animal-associated habitat coral coral metagenome Aruba ?
2 organic material marine biome animal-associated habitat coral coral metagenome US Virgin Islands ?
3 organic material marine biome plant-associated habitat Mangrove Leaf plant metagenome US Virgin Islands ?
4 organic material marine biome coral reef Turf Algae algae metagenome US Virgin Islands ?
5 organic material marine biome animal-associated habitat coral coral metagenome Aruba ?
6 organic material marine biome animal-associated habitat hydrozoans hydrozoan metagenome Aruba ?
7 organic material marine biome animal-associated habitat coral coral metagenome US Virgin Islands ?
8 organic material marine biome animal-associated habitat sponge sponge metagenome US Virgin Islands ?
9 organic material marine biome animal-associated habitat sponge sponge metagenome US Virgin Islands ?
10 organic material marine biome animal-associated habitat coral coral metagenome Aruba ?
11 organic material marine biome coral reef Turf Algae algae metagenome US Virgin Islands ?
12 organic material marine biome animal-associated habitat sponge sponge metagenome US Virgin Islands ?
13 organic material marine biome animal-associated habitat coral coral metagenome US Virgin Islands ?
14 organic material marine biome anthrogenic environmental feature Boat Hull metagenome US Virgin Islands ?
15 organic material marine biome animal-associated habitat sponge sponge metagenome US Virgin Islands ?
16 organic material marine biome animal-associated habitat coral coral metagenome US Virgin Islands ?
17 organic material marine biome animal-associated habitat coral coral metagenome US Virgin Islands ?
18 anthropogenic environmental material urban biome research facility control swab metagenome US Virgin Islands ?
19 anthropogenic environmental material urban biome research facility control swab metagenome US Virgin Islands ?
20 anthropogenic environmental material urban biome research facility control swab metagenome US Virgin Islands ?
21 organic material marine biome plant-associated habitat Mangrove Leaf plant metagenome US Virgin Islands ?
22 organic material marine biome animal-associated habitat coral coral metagenome US Virgin Islands ?
23 organic material marine biome animal-associated habitat coral coral metagenome US Virgin Islands ?
24 organic material marine biome coral reef Turf Algae algae metagenome Aruba ?
25 organic material marine biome animal-associated habitat coral coral metagenome Aruba ?
26 organic material marine biome animal-associated habitat sponge sponge metagenome US Virgin Islands ?

Could you predict the "empo_3" values(which is now in '?') with Animal (saline) or Plant (saline) or Solid (non-saline) or Aqueous (saline) ?

\end{verbatim}
}

\subsection{Zero-shot \textit{E. Coli} risk binary classification prompt example}

\begin{verbatim}
Could you classify the following 56 rows using the Ecoli_binary label, 
where 1 indicates non-safe freshwater and 0 indicates safe freshwater?  

 Date  Lake_Temp_C  Lake_Turb_NTRU  WaveHt_Ft  LL_PreDay  AirportRain48W_in  
2005-05-25         13.3            58.0       1.00     -0.099                0.0
2005-05-26         14.4            11.5       1.00     -0.256                0.0
2005-05-31         17.8             3.2       0.00      0.027                0.1
2005-06-01         17.8             3.5       0.50      0.045                0.0
2005-06-02         17.2             8.4       1.00     -0.062                0.0
2005-06-06         18.3             7.1       0.33      0.184                0.1
2005-06-07         20.0             9.3       0.50     -0.328                0.1
2005-06-08         20.0             3.0       0.00      0.108                0.0
2005-06-09         23.3             1.9       0.00      0.026                0.0
2005-06-13         24.4             2.9       0.00      0.079                0.1
2005-06-14         21.7             5.2       1.00     -0.069                0.3
2005-06-15         21.1            18.5       1.00     -0.148                1.1
2005-06-16         20.0            54.5       1.50      0.184                1.3
2005-06-20         22.2            32.5       0.50     -0.217                0.0
2005-06-21         21.7            26.0       0.50     -0.029                0.0
2005-06-22         22.8            23.0       0.50      0.029                0.2
2005-06-23         22.8            25.5       0.50      0.027                0.1
2005-06-27         24.4             9.8       0.00     -0.075                0.0
2005-06-28         23.9             4.2       0.00      0.049                0.0
2005-06-29         24.4             7.0       0.50      0.072                0.3
2005-06-30         24.4             3.6       0.00     -0.144                0.3
2005-07-05         24.4             6.2       1.00      0.069                0.0
2005-07-06         23.3            16.0       1.50      0.036                0.0
2005-07-07         24.4            12.0       1.00     -0.099                0.0
2005-07-11         24.4             3.8       0.50      0.004                0.0
2005-07-12         24.4             3.0       0.50     -0.073                0.0
2005-07-13         25.6             4.7       0.50     -0.009                0.0
2005-07-14         24.4             4.4       0.50      0.006                0.1
2005-07-18         25.0             4.4       0.50     -0.014                0.2
2005-07-19         24.4             4.3       0.50      0.050                0.4
2005-07-20         26.1             1.7       0.50      0.013                0.2
2005-07-21         23.9             6.1       0.50     -0.138                1.5
2005-07-25         25.6             8.3       0.50     -0.053                0.5
2005-07-26         25.6             2.3       0.50     -0.236                0.3
2005-07-27         24.4            62.5       5.00      0.857                2.9
2005-07-28         22.2            14.7       2.00     -0.584                1.9
2005-08-01         26.1             4.7       0.00     -0.003                0.0
2005-08-02         26.7             3.0       0.50     -0.053                0.0
2005-08-03         27.2             2.0       0.00      0.007                0.0
2005-08-04         26.1             1.6       0.50     -0.089                0.0
2005-08-08         26.7             1.7       1.00     -0.052                0.0
2005-08-09         26.7             1.5       0.00     -0.138                0.0
2005-08-10         26.1             1.9       0.00      0.023                0.0
2005-08-11         26.1             8.0       2.00      0.253                0.3
2005-08-15         25.6            13.5       2.00      0.062                0.3
2005-08-16         24.4             4.8       0.33     -0.095                0.0
2005-08-17         25.6             2.3       0.17     -0.118                0.0
2005-08-18         25.0             2.3       0.66     -0.040                0.0
2005-08-22         24.4            86.5       3.00      0.076                3.6
2005-08-23         24.4            28.5       4.00      0.023                0.1
2005-08-24         24.4            10.2       2.50     -0.050                0.1
2005-08-25         24.4            10.0       2.00     -0.032                0.0
2005-08-29         23.3             2.7       0.50     -0.282                1.2
2005-08-30         23.3             2.4       0.50      0.207                0.0
2005-09-01         22.8            52.0       2.33     -0.167                2.2
2005-09-06         21.7            21.5       1.00     -0.229                0.0

Please show the result in Python list format.

\end{verbatim}

\subsection{Few-shot \textit{E. Coli} risk binary classification prompt example}

\begin{verbatim}


Based on below study:

  Date  Lake_Temp_C  Lake_Turb_NTRU  WaveHt_Ft  LL_PreDay  AirportRain48W_in  Ecoli_binary
2006-06-01         20.0             3.9       0.50      0.040                0.7             0
2006-06-05         18.3            21.8       0.50     -0.010                0.2             0
2006-06-06         18.9            10.0       0.00      0.029                0.0             0
2006-06-07         21.1             3.7       0.00      0.030                0.0             0
2006-06-08         20.0            13.5       0.50      0.010                0.0             1
2006-06-12         18.3            41.1       2.50      0.017                0.0             0
2006-06-13         18.3            27.5       1.50     -0.158                0.0             0
2006-06-14         18.9             6.7       0.50     -0.046                0.0             0
2006-06-15         18.9            22.1       1.00     -0.003                0.0             0
2006-06-19         20.0            11.9       0.50      0.217                1.6             1
2006-06-20         20.6             7.3       1.00      0.092                1.9             0
2006-06-21         21.7             3.8       0.00     -0.325                0.6             0
2006-06-22         21.1            17.2       0.00      0.203                4.6             1
2006-06-23         20.0             5.6       0.00      0.105                3.7             0
2006-06-24         21.1            35.0       2.50     -0.062                0.8             1
2006-06-25         22.8             9.8       0.00      0.148                0.1             1
2006-06-26         21.1            11.2       0.50     -0.017                0.0             0
2006-06-27         21.1            14.5       0.00      0.105                0.7             1
2006-07-01         22.8             5.0       0.00     -0.053                0.0             0
2006-07-02         22.8             5.0       0.00     -0.269                0.0             0
2006-07-03         23.3             5.9       0.50      0.158                0.2             0
2006-07-04         21.7             9.1       0.00      0.095                1.1             0
2006-07-05         21.1            57.5       2.50      0.128                0.6             1
2006-07-11         23.9             7.6       0.50      0.214                0.2             1
2006-07-12         22.8             4.7       0.00     -0.161                1.3             1
2006-07-13         22.2            16.7       1.00      0.272                0.8             1
2006-07-14         22.8             4.2       0.50     -0.085                0.1             0
2006-07-15         23.3            19.4       0.00      0.065                1.0             1
2006-07-16         23.9             4.0       0.00     -0.121                0.5             0
2006-07-18         26.1             5.8       0.50     -0.174                0.0             1
2006-07-22         25.6            23.1       3.00      0.180                0.6             1
2006-07-24         25.6             5.1       0.00     -0.124                0.1             0
2006-07-25         25.0             3.8       0.50     -0.142                0.0             0
2006-07-26         25.6             2.9       0.00      0.119                0.6             0
2006-07-27         24.4             4.4       0.50      0.101                1.1             1
2006-07-28         24.4             9.9       1.00      0.315                2.7             1
2006-07-29         26.1             2.5       0.00     -0.246                1.1             0
2006-07-30         26.7             4.4       0.00      0.063                0.0             0
2006-07-31         26.7             2.7       0.00     -0.112                0.9             0
2006-08-01         27.2             2.0       0.00     -0.023                0.5             0
2006-08-02         27.2             3.2       0.00      0.026                0.0             0
2006-08-03         26.7             7.8       0.50      0.296                0.0             1
2006-08-04         25.6            40.6       3.50     -0.122                0.2             1
2006-08-05         26.1            40.1       3.00     -0.003                0.1             1
2006-08-06         25.6            13.3       0.67     -0.141                0.0             0
2006-08-07         26.7             4.5       0.00     -0.125                0.1             0
2006-08-14         25.0             6.8       0.00     -0.157                0.0             0
2006-08-15         26.1            28.8       2.50      0.101                0.3             1
2006-08-16         25.0             6.9       0.50      0.020                0.1             0
2006-08-17         26.1             5.3       0.33      0.000                0.0             0
2006-08-18         25.6             6.8       0.00     -0.072                0.0             0
2006-08-19         24.4             3.5       0.00      0.062                0.4             0
2006-08-20         23.9            29.8       2.50      0.013                0.5             1
2006-08-21         23.3            32.9       1.50     -0.105                0.1             0
2006-08-22         25.0             9.0       0.00     -0.009                0.0             0
2006-08-23         25.0             6.2       0.50      0.085                0.0             0
2006-08-24         24.4             5.9       0.00      0.006                0.0             0
2006-08-28         24.4            32.2       1.50      0.069                0.4             0
2006-08-29         24.4           258.0       5.00      0.640                2.6             1
2006-08-30         23.3           162.8       2.50     -0.558                1.4             1
2006-08-31         23.3            83.0       4.00      0.229                0.0             1


Could you classify the following 56 rows using the Ecoli_binary label, where 1 indicates non-safe freshwater and 0 indicates safe freshwater?

Date  Lake_Temp_C  Lake_Turb_NTRU  WaveHt_Ft  LL_PreDay  AirportRain48W_in
2005-05-25         13.3            58.0       1.00     -0.099                0.0
2005-05-26         14.4            11.5       1.00     -0.256                0.0
2005-05-31         17.8             3.2       0.00      0.027                0.1
2005-06-01         17.8             3.5       0.50      0.045                0.0
2005-06-02         17.2             8.4       1.00     -0.062                0.0
2005-06-06         18.3             7.1       0.33      0.184                0.1
2005-06-07         20.0             9.3       0.50     -0.328                0.1
2005-06-08         20.0             3.0       0.00      0.108                0.0
2005-06-09         23.3             1.9       0.00      0.026                0.0
2005-06-13         24.4             2.9       0.00      0.079                0.1
2005-06-14         21.7             5.2       1.00     -0.069                0.3
2005-06-15         21.1            18.5       1.00     -0.148                1.1
2005-06-16         20.0            54.5       1.50      0.184                1.3
2005-06-20         22.2            32.5       0.50     -0.217                0.0
2005-06-21         21.7            26.0       0.50     -0.029                0.0
2005-06-22         22.8            23.0       0.50      0.029                0.2
2005-06-23         22.8            25.5       0.50      0.027                0.1
2005-06-27         24.4             9.8       0.00     -0.075                0.0
2005-06-28         23.9             4.2       0.00      0.049                0.0
2005-06-29         24.4             7.0       0.50      0.072                0.3
2005-06-30         24.4             3.6       0.00     -0.144                0.3
2005-07-05         24.4             6.2       1.00      0.069                0.0
2005-07-06         23.3            16.0       1.50      0.036                0.0
2005-07-07         24.4            12.0       1.00     -0.099                0.0
2005-07-11         24.4             3.8       0.50      0.004                0.0
2005-07-12         24.4             3.0       0.50     -0.073                0.0
2005-07-13         25.6             4.7       0.50     -0.009                0.0
2005-07-14         24.4             4.4       0.50      0.006                0.1
2005-07-18         25.0             4.4       0.50     -0.014                0.2
2005-07-19         24.4             4.3       0.50      0.050                0.4
2005-07-20         26.1             1.7       0.50      0.013                0.2
2005-07-21         23.9             6.1       0.50     -0.138                1.5
2005-07-25         25.6             8.3       0.50     -0.053                0.5
2005-07-26         25.6             2.3       0.50     -0.236                0.3
2005-07-27         24.4            62.5       5.00      0.857                2.9
2005-07-28         22.2            14.7       2.00     -0.584                1.9
2005-08-01         26.1             4.7       0.00     -0.003                0.0
2005-08-02         26.7             3.0       0.50     -0.053                0.0
2005-08-03         27.2             2.0       0.00      0.007                0.0
2005-08-04         26.1             1.6       0.50     -0.089                0.0
2005-08-08         26.7             1.7       1.00     -0.052                0.0
2005-08-09         26.7             1.5       0.00     -0.138                0.0
2005-08-10         26.1             1.9       0.00      0.023                0.0
2005-08-11         26.1             8.0       2.00      0.253                0.3
2005-08-15         25.6            13.5       2.00      0.062                0.3
2005-08-16         24.4             4.8       0.33     -0.095                0.0
2005-08-17         25.6             2.3       0.17     -0.118                0.0
2005-08-18         25.0             2.3       0.66     -0.040                0.0
2005-08-22         24.4            86.5       3.00      0.076                3.6
2005-08-23         24.4            28.5       4.00      0.023                0.1
2005-08-24         24.4            10.2       2.50     -0.050                0.1
2005-08-25         24.4            10.0       2.00     -0.032                0.0
2005-08-29         23.3             2.7       0.50     -0.282                1.2
2005-08-30         23.3             2.4       0.50      0.207                0.0
2005-09-01         22.8            52.0       2.33     -0.167                2.2
2005-09-06         21.7            21.5       1.00     -0.229                0.0

Please show the result in Python list format.



\end{verbatim}





\end{document}